\documentclass[letterpaper]{article} 
\usepackage[]{aaai24}  
\usepackage{times}  
\usepackage{helvet}  
\usepackage{courier}  
\usepackage[hyphens]{url}  
\usepackage{graphicx} 
\usepackage{natbib}  
\usepackage{caption} 
\DeclareCaptionStyle{ruled}{labelfont=normalfont,labelsep=colon,strut=off} 

\usepackage{algorithm}
\usepackage{algorithmic}
\usepackage{newfloat}
\usepackage{listings}
\usepackage{bibentry}
\usepackage{amsmath}
\usepackage{amssymb}
\usepackage{booktabs}
\usepackage{subcaption}

\floatstyle{ruled}
\newfloat{listing}{tb}{lst}{}
\floatname{listing}{Listing}
\title{Motion Guided Token Compression for Efficient Masked Video Modeling}
\author {
    Yukun Feng~~~
    Yangming Shi~~~
    Fengze Liu~~~
    Tan Yan}
\affiliations {
    ByteDance \\
    \{yukun.feng1, shiyangming.best, fengze.liu, tan.yan\}@bytedance.com
}

\begin{document}

\maketitle

\begin{abstract}
Recent developments in Transformers have achieved notable strides in enhancing video comprehension. Nonetheless, the O($N^2$) computation complexity associated with attention mechanisms presents substantial computational hurdles when dealing with the high dimensionality of videos. 
This challenge becomes particularly pronounced when striving to increase the frames per second (FPS) to enhance the motion capturing capabilities. 
Such a pursuit is likely to introduce redundancy and exacerbate the existing computational limitations.
In this paper, we initiate by showcasing the enhanced performance achieved through an escalation in the FPS rate. 
Additionally, we present a novel approach, Motion Guided Token Compression (MGTC), to empower Transformer models to utilize a smaller yet more representative set of tokens for comprehensive video representation. Consequently, this yields substantial reductions in computational burden and remains seamlessly adaptable to increased FPS rates.
Specifically, we draw inspiration from video compression algorithms and scrutinize the variance between patches in consecutive video frames across the temporal dimension. 
The tokens exhibiting a disparity below a predetermined threshold are then masked.
Notably, this masking strategy effectively addresses video redundancy while conserving essential information.
Our experiments, conducted on widely examined video recognition datasets, Kinetics-400, UCF101 and HMDB51, demonstrate that elevating the FPS rate results in a significant top-1 accuracy score improvement of over 1.6, 1.6 and 4.0.
By implementing MGTC with the masking ratio of 25\%, we further augment accuracy by 0.1 and simultaneously reduce computational costs by over 31\% on Kinetics-400. 
Even within a fixed computational budget, higher FPS rates paired with MGTC sustain performance gains when compared to lower FPS settings.
\end{abstract}

\section{Introduction}
In recent years, Transformer-based methods have achieved significant improvement in video understanding\cite{Girdhar_2019_CVPR, Sharir2021AnII, Neimark_2021_ICCV, Liu_2022_CVPR, khan2022transformers, wei2022masked, ryali2023hiera,feichtenhofer2022masked,wang2022internvideo,selva2023video}.
The attention mechanism has been demonstrated the effectiveness in building the dependency between visual tokens\cite{chen2020generative,dosovitskiy2021an,Liu2021SwinTH}.
However, it naturally suffers from a $O(N^2)$ computational cost with respect to the input sequence length, which further limits the FPS when we feed video frames.
To tackle this issue, researches try to decompose the spatial and temporal dimension and use two attention modules separately \cite{Arnab_2021_ICCV, bertasius2021spacetime}. 
Such methods can largely reduce the memory cost but it becomes challenging to learn the joint spatial-temporal relationships between video frames, and downgrades the performance eventually.

\begin{figure}[t]
\centering
\includegraphics[width=\linewidth]{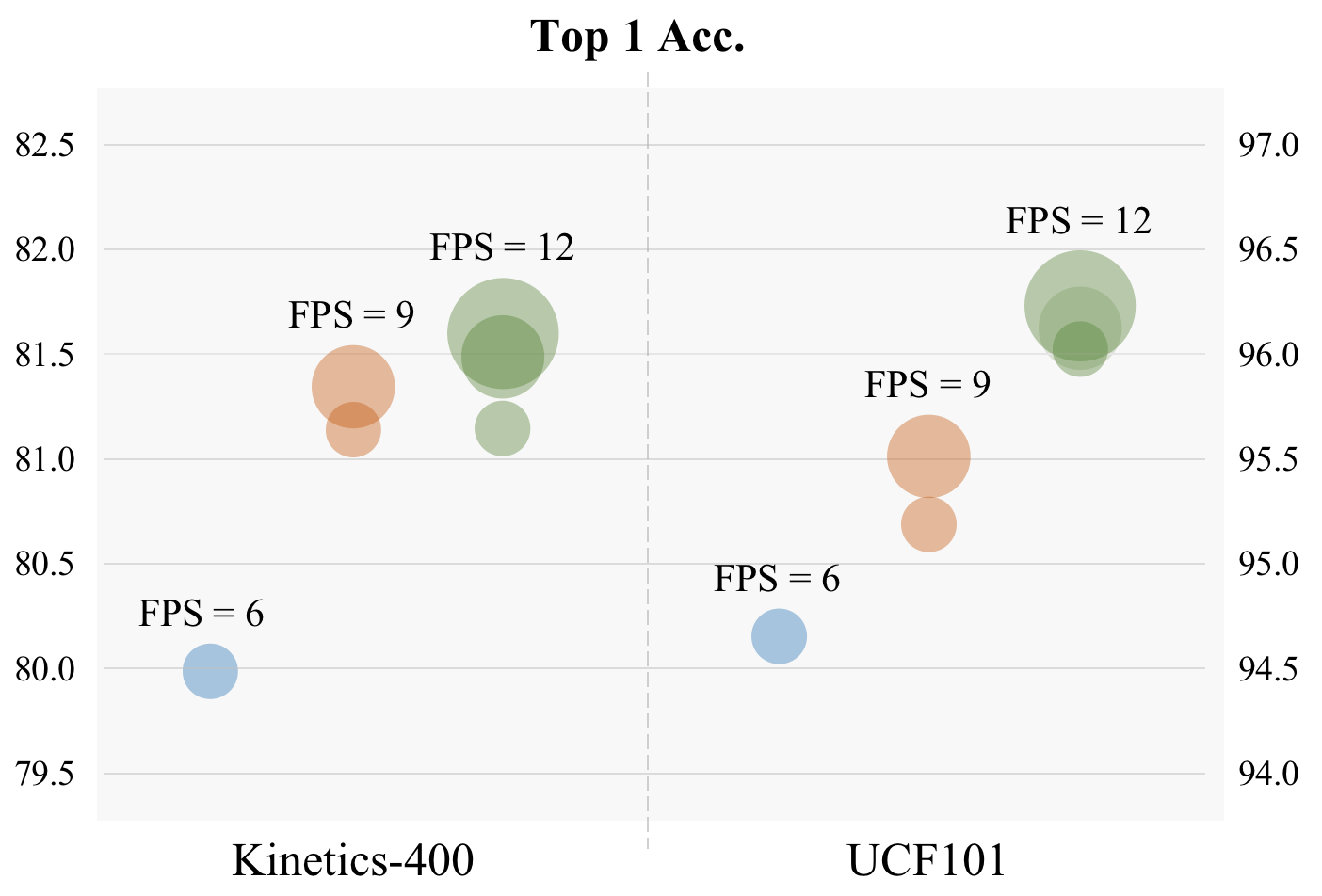}
\caption{\textbf{Overview of the Top-1 Accuracy under different FPS on Kinetics-400 and UCF101.} The relative circle size symbolizes the comparative extent of the computational capacity, which is controlled by the masking ratio and FPS.}
\label{overview}
\end{figure}

Due to such a limitation, most works \cite{selva2023video} fail to incorporate a higher FPS and they normally evaluate the model under a fixed rate.
However, as introduced in \cite{8531714}, a higher FPS allows for more temporal information to be captured, not only by the humans, but also benefit the neural video models with a better representation of video actions. 
This is particularly crucial for tasks that need subtle movements and temporal patterns, such as the action recognition \cite{kay2017kinetics,soomro2012ucf101,goyal2017something,ulhaq2022vision}.
Moreover, a higher FPS is able to reduce the temporal aliasing effect, known as the "strobe effect". 
When the fast-moving objects in low-FPS is likely to appear choppy or blurred, which confuses the video models to understand the content \cite{7406421}. 
Increasing the FPS allows the model to reduce the impact of temporal aliasing, which we believe would result in the better performance for neural video models as well.
Yet, it is acknowledged that increased FPS rates are likely to introduce additional redundancy, and increase the computational cost, particularly given the current attention mechanisms.

In response to the inherent redundancy in videos, video compression algorithms \cite{chen2006introduction} have become a widespread subject of study and application in contemporary video encoding. 
These algorithms primarily concentrate on compressing motion data while maintaining the fidelity of the reconstructed video. 
The key frames and motion vectors extracted from a compressed video format have demonstrated their utility across various visual tasks, as illustrated in \cite{zheng2023dynamic}.
Nonetheless, the potential sparsity of these motion vectors is not fully harnessed, leading to persistent concerns regarding redundancy and computational overhead when integrating compressed videos into computer vision frameworks.

Recently, the application of masked video modeling has demonstrated its effectiveness in the domain of representation learning \cite{tong2022videomae,feichtenhofer2022masked,wang2023videomae}. 
This approach involves a self-supervised task wherein the entire video is reconstructed using a restricted subset of randomly chosen tokens, and yields reasonable reconstruction quality, indicating that spatial-temporal redundancy can be addressed using only a fraction of the video tokens.
However, it's important to note that this masking is typically implemented solely during the pre-training phase and doesn't alleviate the computational constraints during the fine-tuning stage. 
Additionally, the strategy of random token selection still imposes limitations on performance, as each video token shares the same probability of being masked, regardless of its level of informativeness. 
Consequently, essential tokens representing crucial video information might end up being masked.

In an effort to enhance the masking approach, we aim to maintain a high FPS rate while diminishing redundancy and adhering to a predefined computational budget. 
Our inspiration predominantly stems from video compression algorithms, propelling us to introduce Motion Guided Token Compression (MGTC) to address these challenges.
Contrary to the random token masking technique, our approach involves uniformly segmenting the entire video into non-overlapping patches (or cubes, when the temporal block range exceeds 1). 
For each token, we compute the disparity between itself and the token at the corresponding spatial position in subsequent frames. 
This computation signifies the variance in motion across the temporal dimension for the current token. 
Subsequently, we retain tokens with a motion difference surpassing a predefined threshold.
With this method, MGTC guarantees the retention of more informative tokens while effectively discarding redundant tokens through this lightweight token-difference mechanism. 
Importantly, MGTC remains compatible with higher FPS rates, as its masking strategy ensures a consistent number of input tokens.
This approach can be seamlessly integrated into either the evaluation or training phase, effectively alleviating computational constraints. 
As demonstrated in Figure \ref{overview}, we present compelling evidence of the consistent performance enhancement achieved through higher FPS rates within a fixed computational budget, across three extensively studied video datasets, Kinetics-400, UCF101 and HMDB51. 
Additionally, we observe that implementing MGTC with a designated masking ratio further amplifies accuracy while reducing computational costs.
To summarize our contributions:
\begin{enumerate}
    \item[-] We demonstrate the performance gain when increasing the FPS rate. Given a fixed computational budget, a higher FPS is even better with masking 50\% tokens.
    \item[-] We propose the Motion Guided Token Compression strategy, consisting of lightweight motion-guided masking, which is capable of removing spatial-temporal redundancy as well as keeping the action movement, which is able to address the the computational limitation during inference.
    Also, MGTC is able to enhance video representation by applying the masking during training, and further pushes the performance gain.
    \item[-] We find the MGTC is better than other masking methods under different masking ratio. It is even better than using all video tokens when we mask 10-20\% tokens, demonstrating its necessary in removing the video redundancy. 
\end{enumerate}

\section{Related Work}
\paragraph{Video Action Recognition}
Video action recognition is a representative task in video understanding, and it has seen significant advancements in recent years. 
Early research in this field mainly focused on CNN-based approaches, which address the video temporal information with a two-stream CNN network \cite{10.1145/2993148.2997632,Feichtenhofer_2019_ICCV} or the 3D-CNN directly \cite{Cheron_2015_ICCV,10.1145/2671188.2749406,diba2017temporal,Hou_2017_ICCV,BABAEE2018635,8121994,8574597,TU201832,Lin_2019_ICCV,YAO201914,Sudhakaran_2020_CVPR,s20030578,9412193}. 

However, inspired by the success of Transformer in natural language processing \cite{DBLP:journals/corr/VaswaniSPUJGKP17, devlin-etal-2019-bert} and image modeling \cite{dosovitskiy2021an,DBLP:journals/corr/abs-2103-14030} tasks, recent studies \cite{Arnab_2021_ICCV,Liu_2022_CVPR,selva2023video,bertasius2021spacetime,patrick2021keeping} have explored the application of the Transformer architecture to video action recognition.
For instance, ViViT \cite{Arnab_2021_ICCV} and Timesformer \cite{bertasius2021spacetime} have explored factorizing the spatial and temporal dimensions of the input video. 
VideoSwin \cite{Liu_2022_CVPR} and MViT \cite{fan2021multiscale,li2022mvitv2,ryali2023hiera} have proposed hierarchical structures to enable video representation learning and introduce inductive bias to vision transformers. 
Uniformer and its improvements \cite{li2022uniformer,li2022uniformerv2} have designed a hybrid backbone by integrating 3D-CNN to transformers, combining the advantages of both. 
These recent advancements along with other works \cite{yang2022recurring, chen2022adaptformer,ulhaq2022vision,selva2023video} in video action recognition have made great progress and shown promising results.

\begin{figure*}[htbp]
\centering
\includegraphics[width=0.87\linewidth]{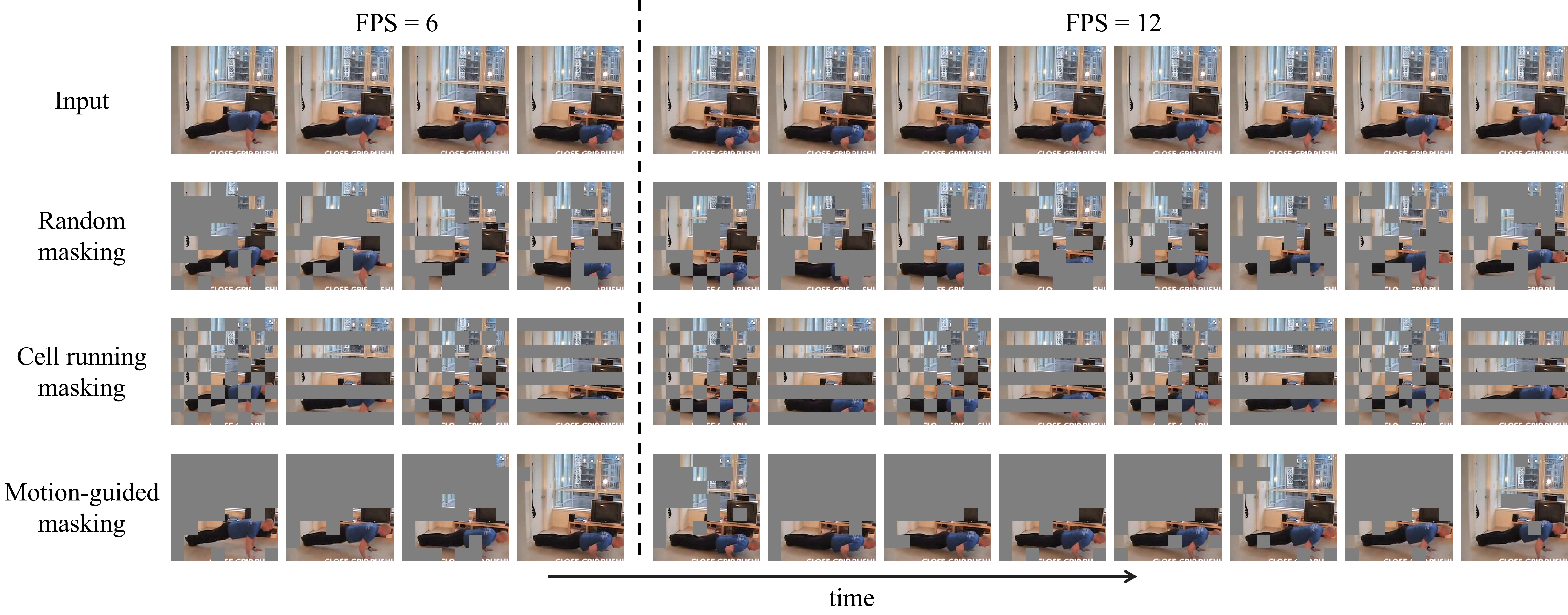}
\caption{\textbf{Comparison between different masking methods, under various FPS rate.} MGTC is able to capture the action movement, and remove the redundant information, especially in higher FPS rate. Here we use a masking ratio of 50\%.}
\label{example-all}
\end{figure*}

\paragraph{Masked Vision Modeling}
The central objective of masked vision modeling is to develop proficient representations of visual data by reconstructing data from deliberately corrupted sources. This concept draws inspiration from the Masked Language Modeling task pioneered by BERT\cite{devlin-etal-2019-bert, liu2019roberta}. 
To achieve the best outcomes, it is essential to design a well-constructed reconstruction task. The reconstructed targets can encompass either the original pixel-level data or derived features, both of which have been extensively explored in recent research \cite{chen2020generative,He_2022_CVPR,bao2022beit,peng2022unified,xie2022simmim,wang2022bevt,girdhar2023omnimae,wang2023masked,bandara2023adamae,sun2023masked}.
The choice of the masking strategy in masked video modeling is of utmost importance. One effective strategy, called "Tube masking," was introduced by VideoMAE\cite{tong2022videomae}, and it has proven superior to other strategies like frame masking and random masking. 
Tube masking prevents information leakage during pre-training, contributing to its success. Additionally, MAR (Motion-AR) \cite{qing2022mar} introduced the concept of Cell Running masking to enhance fine-tuning efficiency, and this concept was further utilized in VideoMAE-v2 \cite{wang2023videomae} for large-scale video pre-training.
Combining these established techniques with advanced video compression algorithms, we introduce a groundbreaking motion-guided masking approach for tokenizing compressed videos. 
This innovative approach promises to improve the efficiency and effectiveness of video tokenization while also minimizing redundancy in the videos and reducing computational demands during the fine-tuning and inference process.

\paragraph{Video Compression}
Video compression is a widely researched area, and numerous works have contributed to the development of efficient and effective video encoding methods \cite{1369699, 4317622, kumar2019novel}. 
One of the seminal works in video compression is the H.264/AVC standard \cite{1369695, zhao2006highly}, which introduced advanced video coding techniques such as motion compensation, transform coding, and entropy coding. 
This standard significantly improved video compression efficiency compared to its predecessors. Another important contribution is the High Efficiency Video Coding (HEVC) standard \cite{6316136}, also known as H.265, which further enhanced compression efficiency by incorporating advanced coding tools like larger block sizes, improved motion compensation, and more sophisticated entropy coding. 

In recent years, there has been a growing interest in deep learning-based video compression methods \cite{ma2019image,NEURIPS2021_96b250a9,9288876}. These approaches leverage neural networks to learn and exploit temporal and spatial redundancies in videos, leading to improved compression performance. Additionally, research efforts have focused on exploring emerging video compression techniques such as perceptual video coding, which takes into account human visual perception to allocate bits more efficiently. Overall, the related works in video compression have made significant strides in achieving higher compression ratios while maintaining acceptable video quality, enabling efficient storage and transmission of video content in various applications.

\section{Methodology}
\label{method}

In order to ensure the compatibility of the video model with an higher FPS rate while concurrently managing feature redundancy within a constrained computational budget, we introduce a straightforward yet efficient masking strategy named Motion Guided Token Compression (MGTC). 
This strategy effectively handles the redundancy by selectively retaining only the informative video patches found between frames.
Notably, MGTC can be seamlessly integrated during the inference stage, maintaining the original computational cost even when the FPS is enhanced. 
Furthermore, the application of MGTC during the training phase could further accelerate the training and decrease the computational cost.

Figure \ref{example-all} shows an example of the comparison of MGTC and other two masking strategies, under a masking ratio of 50\%.
It becomes evident that MGTC not only retains a greater amount of action dynamics but also effectively mitigates feature redundancy across the temporal dimension, especially in a higher FPS setting.

\begin{figure}[t]
\centering
\includegraphics[width=\linewidth]{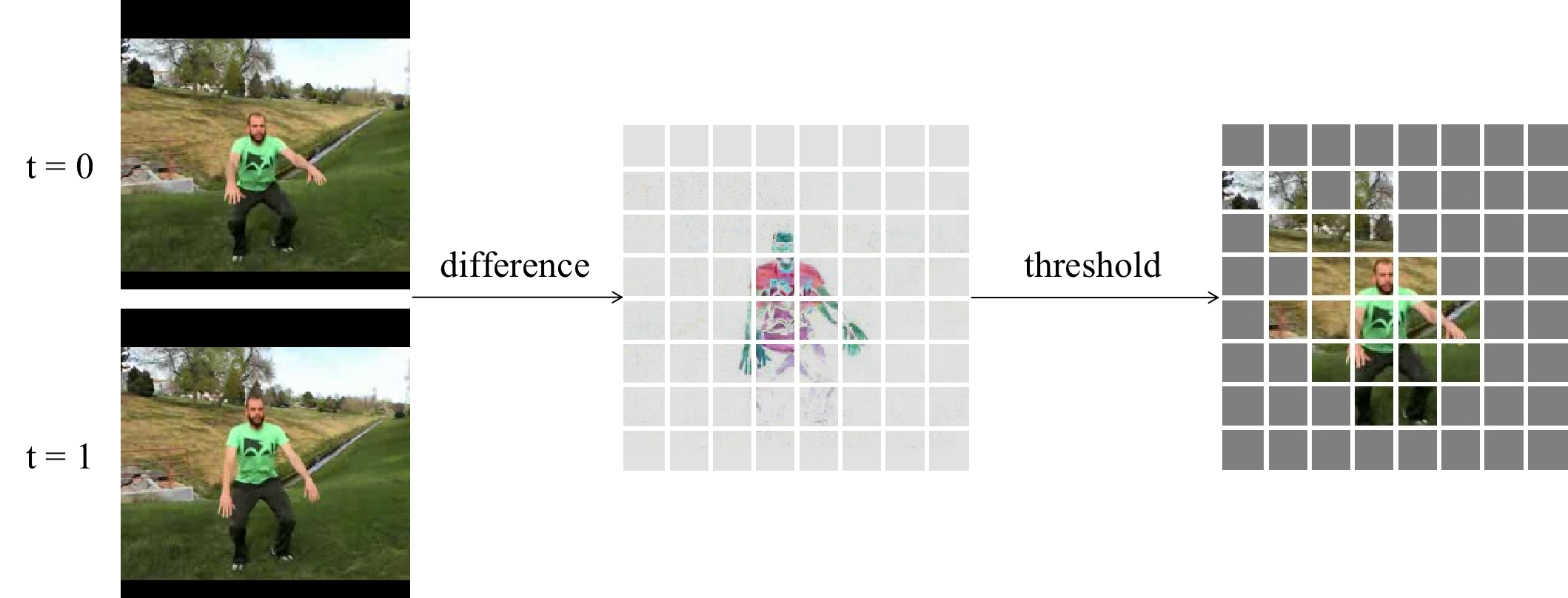}
\caption{\textbf{Workflow of motion-guided masking.} Patch differences are calculated for further masking.}
\label{example}
\end{figure}

\subsection{Motion Guided Token Compression}
The conventional video compression approach, such as H264 \cite{chen2006introduction}, compresses video content by encoding pixel data through the identification and comparison of variations between predicted and observed pixels. 
This process allows for the discrimination between essential and redundant pixel information.

Taking inspiration from the encoding principles of H264, MGTC employs a similar concept to eliminate duplicate video patches between consecutive frames. 
This is achieved through an analysis of pixel disparities across the temporal dimension, encompassing two distinct steps: Sub-block Division and Block Masking.
Figure \ref{example} provides an illustrative instance that shows the MGTC tokenization process, delineating how the frame is partitioned into patches and specifying the patches that are retained.

\paragraph{Sub-block Division}
MGTC first divides the video into multiple non-overlapping sub-blocks.
Given a video with dimensions of $T\times{H}\times{W}$, we use a cube with dimensions of $c\times{p1}\times{p2}$ to tokenize it into a sequence of $L = \frac{T}{c}\times\frac{H}{p1}\times\frac{W}{p2}$ blocks. 
By treating the video as a series of sub-blocks, this sequence effectively retains all the details present in the original video. 
This approach not only improves the ability to capture subtle distinctions in subsequent stages but also seamlessly address the limitation of transformer-based models, thereby creating a practical and feasible input format suitable for diverse applications.

\paragraph{Block Masking}
Since a significant portion of video frames consists of duplicate content, the primary distinction typically revolves around the motion dynamics. 
MGTC efficiently identifies changes in action by examining pixel differences between consecutive blocks within the temporal dimension. 
Specifically, blocks that exhibit minor or no changes in succession are considered redundant and, consequently, are masked from the input sequence.

As we define the above sequence of blocks with a length of $L$, before flattening it into a 1D sequence, we keep its spatial and temporal dimension as $\{\frac{T}{c}\}\times\{\frac{H}{p1}\times\frac{W}{p2}\}$.
Then MGTC compares the pixel difference $D$ with the adjacent available blocks along the temporal dimension.
If $D$ is below a threshold $\lambda$, the blocks will be masked, described as: 
\begin{gather*} 
    Mask_j^i = True~~\text{if}~~D_j^i < \lambda~~\text{else}~~False, \\
    {D}_j^i = MSE(C_j^i,  C_j^{i+1})
\end{gather*} 
where the $i$ and $j$ refer to the index of the temporal and spatial dimension respectively. 
Hence, blocks exhibiting notable pixel alterations are grouped together to effectively represent the motion patterns within the input video. 
For the purpose of representing video appearances, a temporal index is randomly chosen during the training process, similar to a key frame in H264 \cite{chen2006introduction}.
All video patches originating from this selected index are retained.

Within MGTC, the hyperparameter $\lambda$ plays a pivotal role in determining the quantity of cubes necessitating masking. 
Given the inherent diversity in videos, utilizing a uniform threshold across the entire dataset is unfeasible. 
Instead, we dynamically compute the threshold for each specific video based on its unique attributes.
Our approach assumes that a certain proportion of each video constitutes redundancy. Consequently, we set $\lambda$ to match the cube difference value at the upper $N$ percent of all cubes present within that video. 
In the forthcoming analysis section, we will delve into the exploration of varying values for $N$ to assess their impact.

\subsection{Training and Evaluation}
MGTC serves as a versatile plug-and-play enhancement, capable of seamless integration with a diverse range of transformer-based models for video comprehension, whether during training or evaluation.
It's essential to acknowledge that earlier researchers have identified the potency of joint time-space attention across all video cubes. 
However, this approach introduces a computationally intensive complexity of $N^2$. 
Consequently, we opt to utilize VideoMAE as a baseline model, aiming to validate the efficiency and effectiveness of MGTC in this context.

\paragraph{Evaluation with MGTC}
Rather than incorporate MGTC into the training process, we can opt for a simpler yet effective approach. 
We utilize the representations that have been thoroughly trained with all video tokens during fine-tuning. 
In the inference stage, we selectively employ informative tokens that were selected by MGTC.
This method proves to be straightforward, which leverages the advantages of training with the full token set while significantly reducing the number of tokens used during inference. 
It also alleviates the inference bottleneck while still preserving the essential information within the video.

\paragraph{Training with MGTC}
We adhere to the training configuration outlined in VideoMAE \cite{tong2022videomae}, which consists of an initial pre-training phase involving VideoMAE reconstruction, followed by fine-tuning for classification tasks on downstream objectives.
In this study, we do not incorporate MGTC during the pre-training stage. 
Instead, we use the same setup as the original VideoMAE. 
The primary distinction lies in our utilization of a higher FPS during the pre-training phase, leading to an increase in computational demands.
During the fine-tuning phase, we explore two approaches: either using all video tokens, as per the original setting, or applying MGTC to retain informative video cubes that can represent the entire video for training purposes.
We anticipate that integrating further training with MGTC will amplify the representation quality of the chosen video patches, concurrently expediting the training process.

\section{Experiments}

\begin{table*}[htbp]
    \centering
    \begin{subtable}[htbp]{\linewidth}
    \centering
    \begin{tabular}{lccccccc}
\textbf{Model} & \textbf{Backbone} & \textbf{Pre-train} & \textbf{FPS} & \multicolumn{1}{c}{\begin{tabular}[c]{@{}c@{}}\textbf{GFLOPs}\\ (G)\end{tabular}} & \multicolumn{1}{c}{\begin{tabular}[c]{@{}c@{}}\textbf{Param}\\ (M)\end{tabular}} & \multicolumn{1}{c}{\begin{tabular}[c]{@{}c@{}}\textbf{Top-1}\\ (\%)\end{tabular}} & \multicolumn{1}{c}{\begin{tabular}[c]{@{}c@{}}\textbf{Top-5}\\ (\%)\end{tabular}} \\ 
\toprule
SlowFast     & Res101+NL  & IN21K  & 12+3  & $359\times{10}\times{3}$   & 60  & 79.8 & 93.9  \\
Timesformer  & ViT-L      & IN21K   & 6     & $8353\times{1}\times{3}$  & 430 & 80.7 & 94.7 \\
ViViT FE     & ViT-L      & IN21k   & 24    & $3980\times{1}\times{3}$  & 430 & 81.7 & 93.8 \\
Motionformer & ViT-L      & IN21K   & 12    & $1185\times{10}\times{3}$ & 382 & 80.2 & 94.8 \\
VideoSwin    & Swin-L     & IN21K   & 12    & $604\times{4}\times{3}$   & 197 & 83.1 & 95.9 \\
MViTv1       & MViTv1-B   & IN21K       & 12    & $170\times{5}\times{1}$   & 37 & 80.2 & 94.3 \\
BEVT         & Swin-B     & IN-1K+DALLE   & 12    & $282\times{4}\times{3}$   & 88  & 80.6 & N/A  \\
OmniMAE      & ViT-B      & IN-1K+Kinetics-400 & 12    & $180\times{5}\times{3}$   & 87  & 80.6 & N/A  \\
MaskFeat     & MViTv1-L   & Kinetics-400   & 6     & $377\times{10}\times{1}$  & 218 & 84.3 & 96.3 \\
MME      & ViT-B      & Kinetics-400 & 6    & $180\times{7}\times{3}$   & 87  & 81.8 & N/A  \\
$Ada$MAE      & ViT-B      & Kinetics-400 & 6    & $180\times{7}\times{3}$   & 87  & 81.7 & 95.2  \\
$\text{MAR}_{\rho=50\%}$          & ViT-B      & Kinetics-400 & 6  & $86\times{5}\times{3}$    & 94  & 81.0 & 94.4 \\
$\text{MAR}_{\rho=50\%}$          & ViT-L      & Kinetics-400 & 6  & $276\times{5}\times{3}$    & 311  & 85.3 & 96.3 \\
\midrule
$\text{VideoMAE}_{e=800}$ & ViT-B & Kinetics-400 & 6 & $180\times{5}\times{3}$ & 87 & 80.0 & 94.4  \\
$\text{VideoMAE}_{e=800}$ & ViT-B & Kinetics-400 & 9 & $240\times{5}\times{3}$ & 87 & 81.3 & 94.9  \\
$\text{VideoMAE}_{e=800}$ & ViT-B & Kinetics-400 & 12 & $451\times{5}\times{3}$ & 87 & 81.6 & 94.9 \\
\midrule
$\textbf{MGTC}_{\rho=25\%,e=800}$ & ViT-B & Kinetics-400 & 6 & $127\times{5}\times{3}$ & 87 & 80.4 & 94.5  \\
$\textbf{MGTC}_{\rho=10\%,e=800}$ & ViT-B & Kinetics-400 & 9 & $210\times{5}\times{3}$ & 87 & 81.4 & 94.8 \\
$\textbf{MGTC}_{\rho=10\%,e=800}$ & ViT-B & Kinetics-400 & 12 & $392\times{5}\times{3}$ & 87 & 81.8 & 94.9 \\
\midrule
$\text{VideoMAE}_{e=1600}$ & ViT-B & Kinetics-400 & 6 & $180\times{5}\times{3}$ & 87 & 81.5 & 95.0  \\
$\text{VideoMAE}_{e=1600}$ & ViT-B & Kinetics-400  & 12 & $451\times{5}\times{3}$ & 87 &  81.8 & 95.0 \\
$\textbf{MGTC}_{\rho=25\%,e=1600}$ & ViT-B & Kinetics-400 & 6 & $127\times{5}\times{3}$ & 87 & 81.6 & 95.0   \\
$\textbf{MGTC}_{\rho=10\%,e=1600}$ & ViT-B & Kinetics-400  & 12 & $392\times{5}\times{3}$ & 87 & 82.0 & 95.1\\
\midrule
$\text{VideoMAE}_{e=1600}$ & ViT-L & Kinetics-400  & 6 & $597\times{5}\times{3}$ & 305 & 85.2 & 96.8 \\
$\text{VideoMAE}_{e=1600}$ & ViT-L & Kinetics-400  & 12 & $1436\times{5}\times{3}$ & 305 & 85.4 & 97.0 \\
$\textbf{MGTC}_{\rho=50\%,e=1600}$ & ViT-L & Kinetics-400  & 6 & $269\times{5}\times{3}$ & 305 & 85.3 & 97.0 \\
$\textbf{MGTC}_{\rho=25\%,e=1600}$ & ViT-L & Kinetics-400  & 12 & $988\times{5}\times{3}$ & 305 & 85.5 & 96.8 \\
\bottomrule
\end{tabular}
\caption{Comparison with state-of-the-arts on Kinetics-400.}

\end{subtable}
    \begin{subtable}[htbp]{\linewidth}
    \centering
    \begin{tabular}{lccccccc}
\textbf{Model} & \textbf{Backbone} & \textbf{Extra data} & \textbf{FPS} & \multicolumn{1}{c}{\begin{tabular}[c]{@{}c@{}}\textbf{GFLOPs}\\ (G)\end{tabular}} & \multicolumn{1}{c}{\begin{tabular}[c]{@{}c@{}}\textbf{Param}\\ (M)\end{tabular}} & \multicolumn{1}{c}{\begin{tabular}[c]{@{}c@{}}\textbf{UCF101}\\ Top-1(\%)\end{tabular}} & \multicolumn{1}{c}{\begin{tabular}[c]{@{}c@{}}\textbf{HMDB51}\\ Top-1(\%)\end{tabular}}\\ 
\toprule
XDC             & R(2+1)D & IG65M & 12 & N/A & 15 & 94.2 & 67.1  \\
GDT             & R(2+1)D & IG65M & 12 & N/A & 15 & 95.2 & 72.8 \\
CVRL            & Res50   & Kinetics-400  & 12 & N/A & 32 & 92.9 & 67.9 \\
$\text{CORP}_f$ & Res50   & Kinetics-400  & 3  & N/A & 32 & 87.3 & 68.0 \\
$\rho\text{BYOL}$     & Res50   & Kinetics-400  & 3  & N/A & 32 & 94.2 & 72.1 \\
MME             & ViT-B   & Kinetics-400  & $6^*$  & $180\times{5}\times{3}$ & 87 & 96.5 & 78.0 \\ 
\midrule
$\text{VideoMAE}_{e=800}$ & ViT-B & Kinetics-400 & $6^*$ & $180\times{5}\times{3}$ & 87 & 94.6 &  70.1 \\
$\text{VideoMAE}_{e=800}$ & ViT-B & Kinetics-400 & 9  & $240\times{5}\times{3}$ & 87 & 95.5 &   N/A\\
$\text{VideoMAE}_{e=800}$ & ViT-B & Kinetics-400 & $12^*$  & $451\times{5}\times{3}$ & 87 &  96.2 &  74.1\\
\midrule
$\textbf{MGTC}_{\rho=10\%,e=800}$ & ViT-B & Kinetics-400 & $6^*$  & $159\times{5}\times{3}$ & 87 & 94.7 &  70.7\\
$\textbf{MGTC}_{\rho=25\%,e=800}$ & ViT-B & Kinetics-400 & 9  & $167\times{5}\times{3}$ & 87 & 95.2 &  N/A \\
$\textbf{MGTC}_{\rho=50\%,e=800}$ & ViT-B & Kinetics-400 & $12^*$  & $182\times{5}\times{3}$ & 87 & 96.0 &  73.3 \\
\midrule
$\text{VideoMAE}_{e=1600}$ & ViT-B & Kinetics-400 & $6^*$  & $180\times{5}\times{3}$ & 87 &  95.4 &  72.2\\
$\text{VideoMAE}_{e=1600}$ & ViT-B & Kinetics-400  & $12^*$  & $451\times{5}\times{3}$ & 87 & 96.2 &  74.1 \\
$\textbf{MGTC}_{\rho=50\%,e=1600}$ & ViT-B & Kinetics-400 & $6^*$  & $80\times{5}\times{3}$ & 87 &  95.5 & 71.9 \\
$\textbf{MGTC}_{\rho=25\%,e=1600}$ & ViT-B & Kinetics-400  & $12^*$  & $306\times{5}\times{3}$ & 87 & 96.3 &  74.2\\

\bottomrule
\end{tabular}
\caption{Comparison with the state-of-the-arts on UCF101 and HMDB51. The superscript ${}^*$ indicates that it is multiplied by 2 on the HMDB dataset.}
\end{subtable}
\caption{\textbf{System-level comparisons on Kinetics-400, UCF101 and HMDB51 dataset.} The GFLOPs refers to 'FLOPs $\times$ Clips $\times$ Crops' The subscript $\rho$ is the mask ratio during inference, while $e$ is the pre-training epoch.}
\label{main-res}
\end{table*}

\subsection{Datasets}
Our experimentation encompasses two extensively examined video action recognition datasets: Kinetics-400 (Kinetics-400) \cite{kay2017kinetics}, UCF101 \cite{soomro2012ucf101} and HMDB51 \cite{kuehne2011hmdb}.
Kinetics-400 offers a substantial scale with around 240k training videos and 20k validation videos, each spanning 10 seconds in duration. This dataset encompasses a comprehensive range of 400 distinct classes.
In contrast, UCF101 and HMDB51 are more compact datasets, featuring 9.5k/3.5k and 3.5k/1.5k training as well as validation videos.

\subsection{Settings}
In line with the pre-training and fine-tuning configurations established in VideoMAE \cite{tong2022videomae}, we employ the masking strategy outlined in Figure \ref{method} for either the fine-tuning or inference stages.
For the remaining training and evaluation hyperparameters, we preserve the same values as those used in the original VideoMAE.

\begin{figure}[t]
\begin{subfigure}{\linewidth}
\centering
\includegraphics[width=0.9\linewidth]{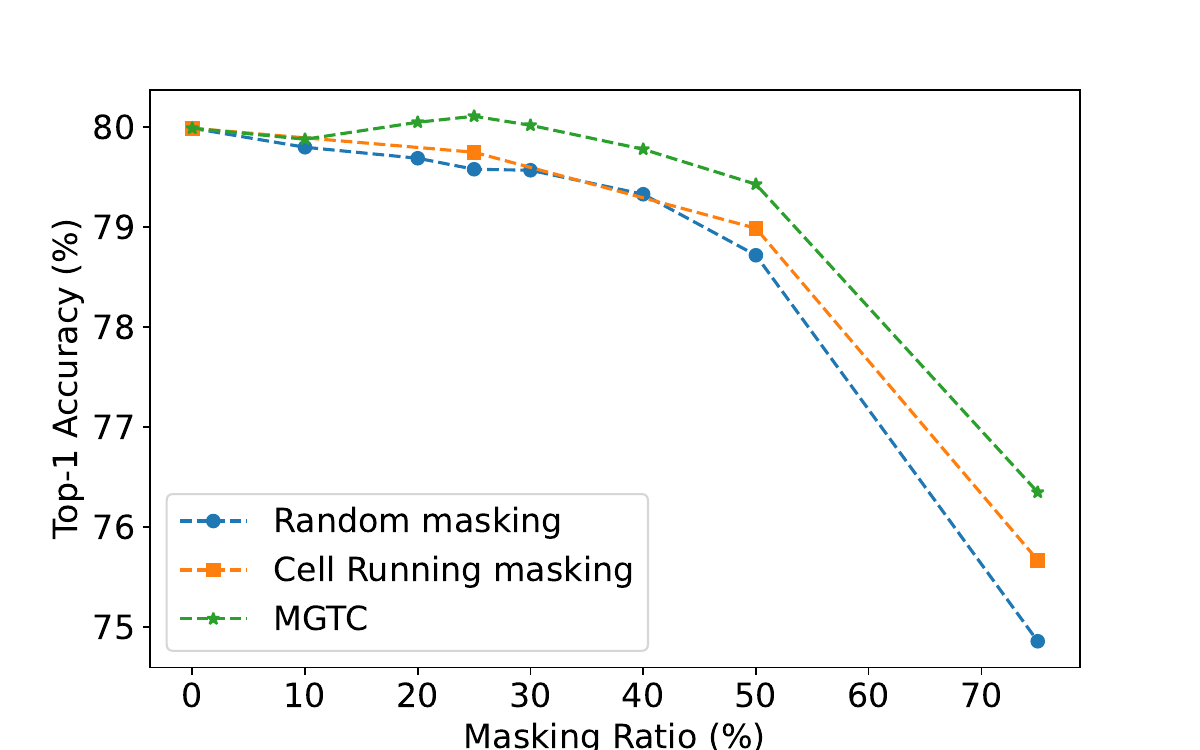}
\caption{Comparison of masking methods under different masking ratios. Experiments are made with 6-FPS setting.
MGTC is consistently better than other Random and Cell Running masking. 
MGTC with 10\% masking even outperforms using all video tokens.}
\label{fig-mask}
\end{subfigure}

\begin{subfigure}{\linewidth}
\centering
\includegraphics[width=0.75\linewidth]{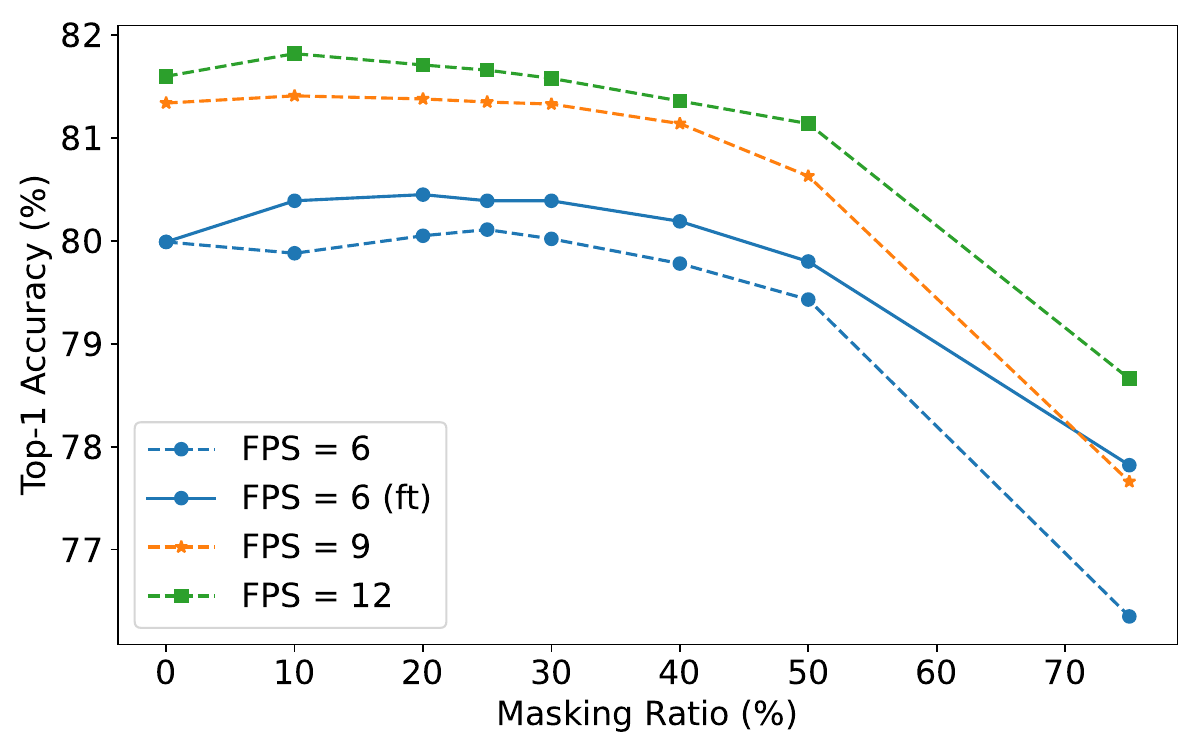}
\caption{Comparison of MGTC in different FPS. A higher FPS brings up performance gain under each masking ratio. "(ft)" refers to training with MGTC.}
\label{fig-mask-fps}
\end{subfigure}
\end{figure}

\subsection{Ablation Study}

We have demonstrated the performance improvement resulting from a higher FPS rate, as highlighted in Figure \ref{overview}. 
Given that incorporating additional video frames significantly escalates computational demands, we have compared three masking strategies aimed at mitigating this limitation.
Remarkably, MGTC consistently outperforms both the Random and Cell Running masking \cite{qing2022mar}. 
Also, surprisingly, the model achieves higher accuracy even when employing a relatively small percentage of masking.
Ablation experiments are conducted with the Kinetics-400 dataset, in which we pre-train the model for 800 epochs while maintaining all other parameters the same.

\paragraph{Masking Methods}
MGTC effectively handles redundancy by masking duplicates while retaining essential tokens, a capability not only found in other methods such as simple random masking and the Cell Running masking \cite{qing2022mar}, but also surpassing them in performance, as illustrated in Figure \ref{fig-mask}. 
This figure clearly demonstrates that MGTC consistently outperforms the other two methods across various masking ratios, highlighting its ability to preserve more representative tokens.
Furthermore, when applying a masking ratio of 10\%-20\%, MGTC achieves superior accuracy scores compared to using all patch tokens, in contrast to the declining scores observed with the other two methods as the masking ratio increases. 
This observation serves as further evidence that MGTC excels at reducing video redundancy, leading to enhanced performance without significant additional computational cost.

\paragraph{Masking under Different FPS}
In Figure \ref{fig-mask-fps}, we present a comparison of MGTC performance across varying FPS settings. 
As the FPS increases, the performance score demonstrates an upward trend and a noteworthy enhancement is observed when transitioning from 6 to 9 FPS, although the relative benefit diminishes when further increasing the FPS to 12.
In settings with higher FPS values, the model reaches its peak relative performance gain with a masking ratio of 10\%, whereas this ratio is 25\% for lower FPS scenarios. 
This consistency suggests that a higher FPS captures more action-related information, resulting in a lower relative redundancy percentage. 
Additionally, it's interesting to note that the relative performance drop in higher FPS settings diminishes when utilizing a masking ratio of 50\%.
This indicates that a higher FPS introduces more redundancy, which, however, is effectively addressed by the MGTC, exerting minimal influence on overall performance.

\paragraph{Masking during Training}
By incorporating MGTC during the training phase, the outcomes depicted in Figure \ref{fig-mask-fps} underscore the effectiveness of training the selected video tokens. 
Training with MGTC enables the model to improve the representation of preserved tokens, resulting in an enhanced video representation and a more substantial performance boost.

\subsection{Main Results}
We compare different FPS settings with either the MGTC or the no-masking strategy in Kinetics-400, UCF101 and HMDB51, the results are shown in Table \ref{main-res}. The SOTA methods we use for comparison are SlowFast \cite{feichtenhofer2019slowfast}, Timesformer \cite{bulat2021spacetime}, ViViT FE \cite{Arnab_2021_ICCV}, MotionFormer \cite{patrick2021keeping}, VideoSwin \cite{liu2021video}, MViTv1 \cite{fan2021multiscale}, BEVT \cite{wang2022bevt}, OmniMAE \cite{girdhar2023omnimae}, MaskFeat \cite{wei2022masked}, MME \cite{sun2023masked}, $ada$MAE \cite{bandara2023adamae}, MAR \cite{qing2022mar}, VideoMAE \cite{tong2022videomae}, XDC \cite{alwassel2020self}, GDT \cite{patrick2020multi}, CVRL \cite{qian2021spatiotemporal}, $\text{CORP}_f$ \cite{hu2021contrast}, $\rho\text{BYOL}$ \cite{feichtenhofer2021large}.

It is noticeable that there is a direct relationship between the FPS and the accuracy scores. 
When the FPS is increased, for all datasets, we consistently observe an improvement in the top-1 accuracy score, from 80.0\%, 94.6\%, 70.1\% to 81.6\%, 96.2\%, 74.1\%, which suggests that optimizing FPS can bring up more motion information, leading to a positive impact on the performance.

Also, when we use a mask ratio of 25\% for 6 FPS and 10\% for the higher FPS setting, there is an improvement up to 0.2 accuracy score. 
Slight masking with the MGTC not only further enhances the performance, but also does so in a more resource-efficient manner, which could be interpreted as removing the video redundancy decrease the noise of input tokens and further improve the accuracy eventually.

Another noteworthy finding from the table is that even when we operate within a fixed budget, the higher FPS settings with MGTC consistently outperform the lower FPS settings. 
This result demonstrates that investing resources in optimizing FPS with our approach is a worthwhile strategy, as it yields better results within the same budgetary limits, making the most of the available resources.

To ensure the scaling ability of the MGTC, we conducted extensive experiments using pre-training on large models, including ViT-Large, over an extended training period of 1600 epochs. 
MGTC consistently produces positive results with a 0.2 gain from higher FPS and an additional 0.1 gain from MGTC. 
In this demanding context, it demonstrates its reliability and potential for broader applicability in other various scenarios.

\begin{figure}[t]
\centering
\includegraphics[width=0.88\linewidth]{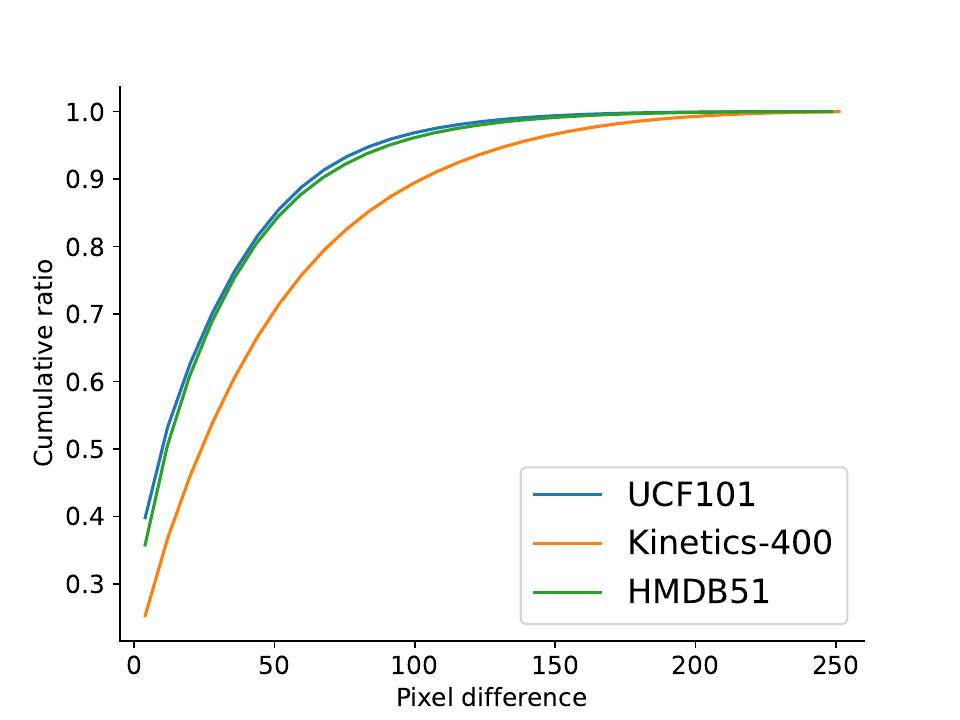}
\caption{Pixel-Residual Distributions under 12 FPS.}
\label{fig-distribution}
\end{figure}
\subsection{Discussions}
\paragraph{Pixel-Residual Distribution}

In this paper, we introduce a video redundancy elimination strategy built upon patch similarity.
For the sake of simplicity, we employ the frame difference as a measure of similarity between patches. 
Our experiments have demonstrated that discarding redundant information not only can diminish computational demands, but also enhance model performance particularly when the mask ratio is lower.

To further underscore the prevalence of excessive redundant data within videos, we have highlighted the frame difference distribution in Figure \ref{fig-distribution}. 
It can be observed that approximately 25\% of frame differences within the Kinetics-400 dataset are near zero, 
and astonishingly, this proportion escalates to 40\% in the case of UCF101. 
Nevertheless, such redundant data can place an undue burden on the model, hence, their removal can contribute to improved modelling performance.
For instance, based on the same model applied to the Kinetics-400 dataset, increasing the mask ratio from 0\% to 40\% improved the accuracy of the model from 80.0\% to a slightly higher 80.4\%.

When adopting a higher FPS and more frames, while also using MGTC to mask certain patches (for instance, 12-FPS, mask ratio=50\%), as opposed to when using 6-FPS without masking, we essentially eliminate low information density data (such as background details) in favor of retaining more high information density data (such as motion content). The result is a substantial boost in overall performance.

\paragraph{Computational Complexity}
We have showcased the efficacy of capturing additional motions through a higher FPS rate. 
Nevertheless, it is essential to acknowledge that this approach also introduces redundancy and imposes supplementary computational burdens.
In this context, we emphasize that MGTC serves to alleviate the augmented computational load by applying masking to a specific percentage of video tokens, as illustrated in Table \ref{main-res}. 
Consequently, the computational expense of employing higher FPS rates with MGTC aligns with that of lower FPS rates, while delivering substantial performance improvements. In essence, MGTC facilitates the maintenance of superior performance within a fixed computational budget for higher FPS settings.

\section{Conclusion}
This paper demonstrates the advantages gained from increasing the FPS rate. 
Additionally, we introduce MGTC as a means to amplify video token representation by eliminating redundant patches, simultaneously reducing computational expenses within a predefined budget. 
Our primary aim is to emphasize the significance of elevating FPS rates in video action recognition, with MGTC representing just one among several potential solutions to tackle computational limitations. 
We encourage future research to dive into the merits of higher FPS settings and to explore more intuitive approaches for alleviating computational constraints.

\clearpage

\bibliography{aaai24}


\end{document}